\documentclass[manuscript,nonacm]{acmart}

\usepackage{mathtools}
\usepackage{booktabs}

\setcopyright{none}
\acmDOI{}
\renewcommand\footnotetextcopyrightpermission[1]{}

\title{RankCert: When Can Simulated Learners Safely Select an AI Tutor? Robust Decision Certification Under Structural Uncertainty}

\author{Nizam Kadir}
\orcid{0000-0002-6725-1133}
\affiliation{%
  \institution{Singapore University of Technology and Design}
  \department{Science, Mathematics and Technology}
  \city{Singapore}
  \country{Singapore}
}
\email{nizam_kadir@mymail.sutd.edu.sg}

\newcommand{\AbstentionMessage}{Insufficient simulation evidence to select a single tutor.}
\newcommand{\abstain}{\bot}
\newcommand{\figurefallback}[1]{%
  \fbox{\parbox[c][1.35in][c]{0.94\linewidth}{%
    \centering Figure file unavailable in this build:\\[2pt]\texttt{#1}}}%
}

\begin{abstract}
Simulation-based tutor selection can be unstable when learner models with adequate
predictive performance imply different policy rankings. RankCert certifies one of
eight equal-budget tutoring policies only when model-averaged utility,
probability-best, posterior regret, cross-domain rank, minimum family coverage, and
leave-one-domain-out and leave-one-visible-family-out averages jointly support the
same candidate; otherwise it abstains. We evaluated RankCert in $1{,}280$ frozen
held-out settings comprising $10{,}240$ decision rows, $163{,}840$ utility
sufficient-statistic rows, and $1{,}280$ immutable raw utility chunks. The design
crossed five rotating held-out oracle families, $64$ scenarios per family, and four
cohort sizes, with $16$ scenarios per stratum and family. Calibration used a
licensed, de-identified EdNet-KT1 derivative containing $5{,}000$ learners and
$590{,}056$ retained responses in learner-disjoint train, validation, and test
partitions; all five family representatives passed the frozen adequacy gate.
H1 was supported: minimum-domain mean pairwise top-1 agreement was $0.272917$
(95\% CI $[0.253646,0.293229]$; Holm $p=0.000400$). H2 was supported: cohort-noise
variance at $n=300$ minus $n=30$ was $-0.000917$ (95\% CI
$[-0.000928,-0.000906]$; Holm $p=0.000250$). H3 was supported: RankCert-minus-point
mean held-out decision loss was $-0.006605$ normalized-outcome units (95\% CI
$[-0.008407,-0.004859]$; Holm $p=0.000250$). H4 was supported: mean within-setting
Kendall $\tau_b$ between response-NLL rank and tutor-ranking agreement was
$0.141644$ (90\% equivalence CI $[0.103632,0.179421]$; Holm $p=0.000250$). H5 was
not supported. At comparable coverage (absolute gap $0.00547$), the
point-certificate-minus-RankCert selective-risk difference was $-0.000213$ (95\%
CI $[-0.003238,0.002384]$; Holm $p=0.929654$). Thus RankCert lowered total decision
loss relative to full-coverage point selection but did not lower conditional regret
relative to the confidence-gated point certificate. ``Safe'' denotes only
benchmark-scoped decision certification under the declared utility and uncertainty
set; no human learning, causal, deployment-effectiveness, or general-safety claim
is made.
\end{abstract}



\begin{document}
\maketitle
\pagestyle{plain}
\thispagestyle{plain}

\section{Introduction}

Simulated learners are often used to compare tutoring policies before a system is
tested or deployed. A simulator provides a compact account of how knowledge,
practice, item difficulty, response history, or instructional thresholds may shape
future performance. That compactness is useful, but it is also a source of decision
risk: two simulator families can both predict observed responses adequately while
implying different rankings of the same tutoring policies. Choosing the policy that
is optimal under a single fitted simulator can therefore convert unresolved model
uncertainty into an unjustifiably definite product decision.

RankCert addresses a narrow question: when does a collection of simulated learners
provide enough evidence to select one AI tutoring policy over its equal-budget
alternatives? The target of inference is the robustness of the selection decision
within a declared benchmark and uncertainty set. It is not a claim that simulated
outcomes equal human learning, that a proxy outcome is validated against a human
endpoint, or that any selected tutor is effective in deployment.

Throughout this article, ``safe'' has one restricted meaning: a policy decision
meets the frozen RankCert certificate inside this benchmark under the declared
utility and predictive-adequate uncertainty set. The term does not denote general
system safety, causal benefit, fairness, security, or suitability for deployment.

RankCert treats structural uncertainty as a first-class part of the
decision. Five learner-model families represent distinct mechanisms: a
Bayesian-knowledge-tracing transition model, a performance-factors-analysis
practice model, a dynamic item-response model, an attentive-history model, and a
causal-threshold model. The causal-threshold family is a mechanism-shaped
predictive surrogate; its name and structure do not identify or estimate a causal
effect. A family enters a decision-specific uncertainty set only if
it has passed the calibration study's predictive-adequacy checks, whose family-level
flags are fixed before confirmatory evaluation. RankCert
then asks whether the same policy remains preferable when evidence is aggregated
across the adequate families and stress-tested across domains.

Certification is deliberately more demanding than point selection. A candidate
must maximize model-averaged utility, have probability of being best of at least
$0.50$, have posterior $95$th-percentile regret no greater than $0.05$, hold the
posterior-mean top rank in at least three of four domains, retain at least three
predictive-adequate simulator families, survive every leave-one-domain-out average,
and survive every leave-one-visible-family-out average. Failure of any condition
produces the fixed message:
\emph{\AbstentionMessage}

The evaluation exposed rather than concealed disagreement. A setting
is defined by its held-out simulator family, scenario draw and stratum, and cohort
size. Every setting contains all eight fixed, equal-budget tutoring-policy profiles
and all four neutral domains. Cohort sizes are $30$, $60$, $120$, and $300$. In
each rotating fold, four families are visible to the decision rules and the fifth
defines the held-out oracle. Scenarios are stable, ambiguous,
structural-conflict, or misspecified. Licensed, de-identified EdNet-KT1 secondary data were
used for calibration and predictive checks; all decision outcomes were evaluated in
the controlled benchmark. No participants are recruited and no learner receives an
intervention as part of this study.

This article contributes (i) a formal decision target that distinguishes robust
policy certification from simulator fit, (ii) an abstaining certificate with
explicit structural and cross-domain gates, and (iii) a confirmatory evaluation
over $1{,}280$ held-out settings. RankCert reduced overall decision loss relative
to the full-coverage point selector, while its selective risk did not differ from
the comparable confidence-gated point certificate. The latter null result is a
central boundary on the empirical claim.

\section{Related Work}

\subsection{Learner Models and Knowledge Tracing}

Learner simulators encode different assumptions about latent knowledge, practice,
item properties, temporal change, memory, and treatment response. Transition-based
knowledge tracing emphasizes latent mastery dynamics \cite{corbett1995};
practice-count models emphasize accumulated opportunities \cite{pavlik2009}; deep
and attentive models learn flexible dependence on prior interactions
\cite{piech2015,ghosh2020}; and broader knowledge-tracing work includes dynamic and
item-sensitive representations \cite{abdelrahman2023,liu2022}. The
mechanism-shaped causal-threshold surrogate used here represents discontinuous
predictive responses to instruction but estimates no causal effect. Predictive
similarity among such models does not imply
agreement about which sequential policy should be chosen, because policy evaluation
depends on counterfactual trajectories rather than one-step prediction alone.

\subsection{Model Uncertainty and Robust Decisions}

Model averaging, minimax regret, lower-tail optimization, and distributionally
robust decision-making provide different responses to uncertainty. Averaging can
reduce sensitivity to a single model but may hide consequential disagreement;
worst-case methods protect against adverse models but can be overly conservative;
and posterior decision rules depend on both the utility model and the uncertainty
representation \cite{hansen2007,nilim2005}. More generally, predictive success can
coexist with consequential modeling differences or underspecification
\cite{breiman2001,damour2022}. RankCert combines model averaging with explicit
evidence gates and an option to make no unique selection.

Decision-focused learning and predict-then-optimize methods connect predictive
model training to downstream decision quality \cite{wilder2019,elmachtoub2022}.
Computer-model calibration addresses a different uncertainty problem by relating a
simulator, its parameters, discrepancy, and observations \cite{kennedy2001}. These
lines of work are relevant context rather than identical baselines. RankCert does
not learn a decision-aware predictor; it evaluates an abstaining policy-selection
certificate under a frozen ensemble of structurally distinct learner simulators.

\subsection{Selective Prediction, Abstention, and Certification}

Selective systems trade coverage for reliability by withholding a prediction or
decision when evidence is weak \cite{geifman2017,geifman2019}. In a
policy-selection setting, abstention must be
defined at the decision level: the procedure declines to name a single tutor rather
than merely attaching uncertainty to a ranking. A useful certificate must also make
its scope explicit. RankCert certifies a choice relative to a benchmark, utility,
and adequate-family set; it does not certify general safety, fairness, or human
learning benefit.

\subsection{Simulation-Based Evaluation of Adaptive Systems}

Simulation benchmarks enable controlled variation of sample size, structural
agreement, and misspecification. Their internal control does not by itself establish
external validity. The present design therefore uses secondary data to constrain
calibration while preserving a synthetic held-out oracle for decision-loss
measurement. This division supports a test of decision robustness without treating
the benchmark as a substitute for human evaluation.

Reinforcement learning for instructional sequencing studies how tutoring actions
can be organized as a learned policy problem \cite{doroudi2019}, while offline
policy evaluation has been developed for educational games, course recommendation,
and curriculum design using logged behavior \cite{mandel2014,hoiles2016}. These
approaches are not treated as interchangeable baselines for RankCert. RankCert's
distinct focus is structural simulator disagreement, held-out-family evaluation,
and abstention rather than estimating the value of one learned policy from logged
behavior.

\section{Problem Formulation}

\subsection{Policies, Domains, and Simulator Families}

Let $\mathcal{P}=\{p_1,\ldots,p_8\}$ be the fixed set of equal-budget tutoring-policy
profiles, $\mathcal{D}=\{d_1,\ldots,d_4\}$ the neutral benchmark domains, and
$\mathcal{M}=\{m_1,\ldots,m_5\}$ the simulator families. The five elements of
$\mathcal{M}$ are, respectively, BKT-transition, PFA-practice, dynamic IRT,
attentive history, and causal threshold. ``Equal budget'' means that the resource
constraint used to expose or allocate tutoring actions is identical across policy
profiles within a held-out setting. The exact profiles and budget accounting were
frozen before confirmatory execution and were not changed in response to pilot
rankings.

A held-out setting $z=(h,s,n)$ is defined by held-out simulator family $h$, scenario
draw $s$ with its predeclared stratum, and cohort size $n$. The scenario draw carries
the residual parameter and cohort realization used for that setting; a domain is not a separate primary
decision unit. Every setting contains all policies in $\mathcal{P}$ and all domains
in $\mathcal{D}$. The visible-family pool is
\begin{equation}
  \mathcal{V}_z=\mathcal{M}\setminus\{h\},
  \qquad |\mathcal{V}_z|=4,
  \label{eq:visible-set}
\end{equation}
and the held-out family supplies the oracle. Rotation over $h\in\mathcal{M}$ makes
each family the oracle in turn.

For policy $p$, visible family $m$, domain $d$, and setting $z$, let $U_{pmdz}$
denote delayed utility on the predeclared common scale, with larger values
preferred. Let $U^{\mathrm{orc}}_{pdz}$ denote the corresponding held-out-family
oracle utility and
\begin{equation}
  U^{\mathrm{orc}}_{pz}=\frac{1}{|\mathcal{D}|}
  \sum_{d\in\mathcal{D}}U^{\mathrm{orc}}_{pdz}
  \label{eq:oracle-average}
\end{equation}
its four-domain setting average. Oracle utilities are available only for held-out
evaluation; no decision rule may use them when selecting or abstaining.

\subsection{Predictive-Adequacy Uncertainty Set}

Let $A_m^{\mathrm{cal}}\in\{0,1\}$ be the pass/fail flag assigned to family $m$ by
the separate calibration study. These flags are fixed before benchmark evaluation
and do not vary by setting, domain omission, scenario draw, or cohort size. The
operative uncertainty set for setting $z$ is the predictive-adequate subset of its
four visible families,
\begin{equation}
  \mathcal{A}_z=\{m\in\mathcal{V}_z:A_m^{\mathrm{cal}}=1\},
  \label{eq:adequate-set}
\end{equation}
where the held-out oracle is never available to the decision rule. Predictive
diagnostics, aggregation rules, and pass/fail cutoffs were fixed before the
confirmatory run. A family that fails does not
contribute utility draws, probability-best calculations, regret quantiles, or the
family-count gate. This restriction prevents poor predictive fit from being treated
as useful diversity while retaining structurally different models that are
empirically plausible.

For adequate families, RankCert uses an equal-family model average so that a family
with more parameterizations or posterior samples does not receive greater weight
merely because of its representation. The posterior mean model-averaged utility in
domain $d$ and its setting-level four-domain average are
\begin{equation}
  \bar U_{pdz}=
  \frac{1}{|\mathcal{A}_z|}
  \sum_{m\in\mathcal{A}_z}
  \mathbb{E}\!\left[U_{pmdz}\mid \mathcal{I}_z,m\right],
  \qquad
  \bar U_{pz}=\frac{1}{|\mathcal{D}|}
  \sum_{d\in\mathcal{D}}\bar U_{pdz},
  \label{eq:model-average}
\end{equation}
where $\mathcal{I}_z$ is the information available to the decision rule. If
$\mathcal{A}_z$ is empty, the procedure must abstain.

\subsection{Probability of Being Best and Regret}

For posterior draw $b$, define the equal-family, equal-domain utility draw
\begin{equation}
  U^{(b)}_{pz}=\frac{1}{|\mathcal{D}|\,|\mathcal{A}_z|}
  \sum_{d\in\mathcal{D}}\sum_{m\in\mathcal{A}_z}U^{(b)}_{pmdz}
  \label{eq:utility-draw}
\end{equation}
and let $\mathcal{T}^{(b)}_z=\operatorname*{arg\,max}_{q\in\mathcal{P}}
U^{(b)}_{qz}$ be the tied top set for that draw. Probability-best mass is split
equally among all policies tied in a draw:
\begin{equation}
  \pi_{pz}=\mathbb{E}_b\!\left[
  \frac{\mathbf{1}\{p\in\mathcal{T}^{(b)}_z\}}
  {|\mathcal{T}^{(b)}_z|}\right].
  \label{eq:prob-best}
\end{equation}
Posterior regret for selecting $p$ is
\begin{equation}
  G^{(b)}_{pz}=\max_{q\in\mathcal{P}}U^{(b)}_{qz}-U^{(b)}_{pz},
  \qquad
  g^{.95}_{pz}=Q_{.95}\!\left(G^{(b)}_{pz}\mid
  \mathcal{I}_z,\mathcal{A}_z\right).
  \label{eq:posterior-regret}
\end{equation}

Within domain $d$, define the posterior-mean rank
\begin{equation}
  r_{pdz}=\operatorname{rank}_{q\in\mathcal{P}}
  \left(-\bar U_{qdz}\right),
  \label{eq:domain-rank}
\end{equation}
so that rank one is best. Rank ties are retained; no ad hoc ordering is imposed.

\subsection{Held-Out Decision Loss}

For a method $a$ and held-out setting $z$, let $\delta_a(z)\in
\mathcal{P}\cup\{\abstain\}$ be its action. Decision loss is
\begin{equation}
L_a(z)=
\begin{cases}
\displaystyle
\max_{q\in\mathcal{P}} U^{\mathrm{orc}}_{qz}
-U^{\mathrm{orc}}_{\delta_a(z),z},
& \delta_a(z)\in\mathcal{P},\\[6pt]
0.01, & \delta_a(z)=\abstain.
\end{cases}
\label{eq:decision-loss}
\end{equation}
Thus a selection is charged its oracle regret, whereas abstention has the fixed
cost $0.01$. This endpoint evaluates the decision made after simulation; it does
not estimate a human treatment effect.

\section{RankCert Decision Procedure}

For setting $z$, RankCert first computes $\mathcal{A}_z$ and the quantities in
Eqs.~\eqref{eq:model-average}--\eqref{eq:domain-rank}. Let
\begin{equation}
  p_z^*=\operatorname*{arg\,max}_{p\in\mathcal{P}}\bar U_{pz}.
  \label{eq:candidate}
\end{equation}
If the model-averaged maximizer is not unique, RankCert abstains; it does not apply
a tie-breaker. Otherwise, the candidate is certified only if every condition below
holds.

\begin{enumerate}
  \item \textbf{Model-averaged winner:} $p_z^*$ is the unique maximizer in
  Eq.~\eqref{eq:candidate}.
  \item \textbf{Probability-best gate:} $\pi_{p_z^*z}\geq 0.50$.
  \item \textbf{Regret gate:} $g^{.95}_{p_z^*z}\leq 0.05$.
  \item \textbf{Cross-domain rank gate:}
  $\sum_{d\in\mathcal{D}}\mathbf{1}\{r_{p_z^*dz}=1\}\geq 3$.
  \item \textbf{Structural coverage gate:} $|\mathcal{A}_z|\geq 3$.
  \item \textbf{Leave-one-domain-out average survival:} for every
  $d_0\in\mathcal{D}$, the same $p_z^*$ is the unique maximizer of
  \begin{equation}
    \bar U^{(-d_0)}_{pz}=
    \frac{1}{(|\mathcal{D}|-1)|\mathcal{A}_z|}
    \sum_{d\in\mathcal{D}\setminus\{d_0\}}
    \sum_{m\in\mathcal{A}_z}
    \mathbb{E}\!\left[U_{pmdz}\mid\mathcal{I}_z,m\right].
    \label{eq:lodo-average}
  \end{equation}
  The calibration-study adequacy flags and $\mathcal{A}_z$ are held fixed when a
  domain is omitted.
  \item \textbf{Leave-one-visible-family-out average survival:} for every
  $m_0\in\mathcal{V}_z$, the same $p_z^*$ is the unique maximizer of
  \begin{equation}
    \bar U^{(-m_0)}_{pz}=
    \frac{1}{|\mathcal{D}|\,|\mathcal{A}_z\setminus\{m_0\}|}
    \sum_{d\in\mathcal{D}}
    \sum_{m\in\mathcal{A}_z\setminus\{m_0\}}
    \mathbb{E}\!\left[U_{pmdz}\mid\mathcal{I}_z,m\right].
    \label{eq:lofo-average}
  \end{equation}
  Omitting a visible family removes it from the average without refitting or
  reclassifying any family. If $m_0$ was not predictive-adequate, the average is
  unchanged. This survival check was implemented and predeclared.
\end{enumerate}

The decision rule is therefore
\begin{equation}
\delta_{\mathrm{RC}}(z)=
\begin{cases}
p_z^*, & \text{if all certification conditions hold},\\
\abstain, & \text{otherwise}.
\end{cases}
\label{eq:rankcert-rule}
\end{equation}
Whenever $\delta_{\mathrm{RC}}(z)=\abstain$, the system output must be exactly:
\emph{\AbstentionMessage} No alternate policy, softened recommendation, or
unregistered tie-breaker may be substituted after abstention.

\section{Experimental Design}

\subsection{Benchmark Factors}

The confirmatory benchmark used the fully crossed design in
Table~\ref{tab:design}. Confirmatory replicate count, random seeds, policy-profile
specifications, utility construction, predictive-adequacy thresholds, and
comparator tuning constants were fixed in a time-stamped specification before
results were generated.

\begin{table}[t]
  \caption{Predeclared benchmark structure. Each held-out setting contains every
  domain and policy. Labels identify design cells rather than empirical findings.}
  \label{tab:design}
  \centering
  \begin{tabular}{@{}ll@{}}
    \toprule
    Factor & Levels \\
    \midrule
    Setting index & Held-out family; scenario draw/stratum; cohort size \\
    Within each setting & Eight equal-budget policies; four neutral domains \\
    Family rotation & Four visible families; fifth family as held-out oracle \\
    Simulator families & BKT-transition; PFA-practice; dynamic IRT; \\
    & attentive history; causal threshold \\
    Cohort size & $30$; $60$; $120$; $300$ \\
    Scenario stratum & Stable; ambiguous; structural conflict; misspecified \\
    \bottomrule
  \end{tabular}
\end{table}

The policy profiles were treated as fixed decision alternatives, not as named
pedagogical personas. Each received the same budget within a setting. Their complete
machine-readable definitions were versioned before the confirmatory run, and no
profile was revised after its relative performance was observed.
All four domains occur together in every setting and contribute to the single
setting-level decision. They use neutral identifiers and common outcome semantics;
domain labels must not imply unmeasured subject-specific or demographic
generalization.

\subsection{Scenario Strata}

The \emph{stable} stratum represents cells in which adequate structural families
induce a clear and persistent ordering. The \emph{ambiguous} stratum represents
near-indifference or posterior overlap among leading policies. The
\emph{structural-conflict} stratum represents cells in which predictive-adequate
families favor different policies because of their structural assumptions. The
\emph{misspecified} stratum represents held-out mechanisms or perturbations not
well represented by the candidate uncertainty set. These descriptions defined the
stress conditions; their operational generators and assignment rules were fixed
without inspecting confirmatory method comparisons.

\subsection{Leave-One-Family-Out Evaluation}

Each of the five simulator families is held out in turn. The held-out mechanism
generates or defines the evaluation environment and oracle utilities, while the
remaining four families are visible to the decision rules. The visible-family pool
is filtered by the calibration-study pass/fail flags; those flags are not estimated
again within a benchmark setting.
The held-out family and oracle outcomes were inaccessible to every selector. The
misspecified stratum used predeclared out-of-set perturbations generated
independently of method outcomes.
All methods receive the same visible families, training information, policy
budgets, posterior draw budget, and held-out settings. Each method issues at most
one policy selection per setting, using all four domains available to that setting.

\subsection{Comparators}

RankCert is compared with seven predeclared alternatives.

\begin{enumerate}
  \item The \textbf{single-simulator point selector} chooses the policy with the
  largest posterior-mean utility under one predeclared simulator and never uses
  structural disagreement as an abstention reason.
  \item The \textbf{confidence-gated point certificate} applies a predeclared
  within-simulator confidence gate to the point selector and abstains when that gate
  fails.
  \item The \textbf{equal-weight model average} selects the policy with largest
  mean utility across predictive-adequate families without RankCert's additional
  certification gates.
  \item \textbf{Probability-best thresholding} selects the policy with largest
  probability of being best only when its predeclared probability threshold is met;
  otherwise it abstains.
  \item \textbf{Top-set selection} forms a simultaneous top set and names a single
  policy only when that set is a singleton; a non-singleton set maps to abstention
  for decision-loss scoring.
  \item \textbf{Minimax regret} selects the policy minimizing worst-family
  posterior expected regret over the predictive-adequate set.
  \item \textbf{Lower-tail distributionally robust selection} chooses the policy
  maximizing a predeclared lower-tail utility criterion over admissible family
  weights.
\end{enumerate}

Comparator thresholds, tail levels, and uncertainty radii were frozen before
confirmatory execution and were not optimized on held-out decision loss. Ties are
handled only as specified by the implemented,
predeclared comparator definitions; no outcome-dependent or ad hoc tie-breaker may
be added. When a comparator cannot act because no required family is adequate, it
abstains and receives the same abstention cost.

\subsection{Development and Confirmatory Separation}

Pilot runs were used only to verify code paths, numerical stability, runtime, and
executability. Pilot outputs did not determine hypotheses, thresholds, scenario
boundaries, reported effect estimates, or Results prose. Confirmatory outputs were
written separately with an immutable manifest linking code version, inputs, seeds,
exclusions, and analysis outputs. No pilot value is reported as a finding.

\section{Public-Data Calibration}

The empirical calibration source is the licensed, de-identified EdNet-KT1
secondary dataset. It is used to constrain plausible response dynamics, parameter
ranges, and predictive checks for simulator families. It is not used to claim that
the benchmark reproduces human learning, that any benchmark utility is a validated
human endpoint, or that a selected policy improves outcomes for EdNet learners
\cite{choi2020}. The confirmatory derivative contained $5{,}000$ learners and
$590{,}056$ retained responses. The learner-disjoint split contained $3{,}533$
training learners with $402{,}650$ responses, $760$ validation learners with
$82{,}332$ responses, and $707$ test learners with $105{,}074$ responses.
Within-learner temporal order was preserved for sequential prediction. Processing
was limited to licensed fields required by the frozen calibration specification.

Predictive adequacy was assigned from validation data only and independently of
downstream policy ranking. The frozen gate was the conjunction of
\begin{equation}
  \mathrm{AUC}\geq 0.55,\qquad
  \mathrm{NLL}_{\mathrm{baseline}}-\mathrm{NLL}_{m}\geq 0.002,
  \qquad \mathrm{ECE}\leq 0.08,
  \label{eq:adequacy-gate}
\end{equation}
where the validation baseline NLL was $0.655201$. Across the five representatives,
validation AUC ranged from $0.631241$ to $0.726513$, validation NLL from $0.580483$
to $0.630700$ (improvements from $0.024501$ to $0.074718$), and validation ECE from
$0.008998$ to $0.020503$. All five representatives passed this conjunction before
test metrics were examined. The resulting family-level flags were then held fixed
for the entire synthetic benchmark, including every domain omission and scenario
draw. The test partition was reserved for reporting and never contributed to the
adequacy decision. Passing indicates only predictive adequacy for inclusion when a
family is visible; it does not establish causal correctness.

\section{Analysis}

\subsection{Primary Estimand}

Let $a_0$ denote the single-simulator point selector. For the set $\mathcal{Z}$ of
held-out confirmatory settings, the primary estimand is the paired mean decision-loss
difference
\begin{equation}
  \Delta_{\mathrm{RC},0}
  =\frac{1}{|\mathcal{Z}|}\sum_{z\in\mathcal{Z}}
  \left[L_{\mathrm{RC}}(z)-L_{a_0}(z)\right].
  \label{eq:primary-estimand}
\end{equation}
A negative value favors RankCert. Pairing is maintained at the complete held-out
setting, so both methods face the same held-out oracle family, scenario draw and
stratum, cohort size, four domains, eight policies, and random environment.

\subsection{Secondary Metrics}

The predeclared secondary metrics are coverage, selective risk, conditional
wrong-selection probability, top-1 agreement, Kendall $\tau_b$, and variance
attributable to simulator family. Coverage is
\begin{equation}
  C_a=\frac{1}{|\mathcal{Z}|}\sum_{z\in\mathcal{Z}}
  \mathbf{1}\{\delta_a(z)\neq\abstain\}.
  \label{eq:coverage}
\end{equation}
Selective risk is mean held-out oracle regret conditional on selection,
\begin{equation}
  R_a^{\mathrm{sel}}=
  \frac{\sum_{z\in\mathcal{Z}}\mathbf{1}\{\delta_a(z)\neq\abstain\}
  \left[\max_{q\in\mathcal{P}}U^{\mathrm{orc}}_{qz}
  -U^{\mathrm{orc}}_{\delta_a(z),z}\right]}
  {\sum_{z\in\mathcal{Z}}\mathbf{1}\{\delta_a(z)\neq\abstain\}},
  \label{eq:selective-risk}
\end{equation}
when coverage is nonzero. Conditional wrong-selection probability is the
probability, among selections, that the selected policy is not in the held-out
oracle top set. Top-1 agreement records agreement in the top-ranked tutor policy
across simulator families. Kendall $\tau_b$ measures tutor-ranking concordance
while retaining ties. Variance attributable to simulator family is obtained from
the predeclared delayed-utility decomposition below.

\subsection{Confirmatory Hypotheses}

H1 and H2 characterize the structural decision problem; H3 and H5 evaluate
RankCert; and H4 prevents predictive fit from being treated as a surrogate for
decision validity. The five predeclared hypotheses are:

\begin{description}
  \item[H1:] Top-1 tutor-policy agreement across simulator families is materially
  below perfect agreement in at least one predeclared domain.
  \item[H2:] Increasing simulated cohort size reduces within-simulator cohort
  variance but does not eliminate between-family structural variance.
  \item[H3:] At abstention cost $0.01$ normalized-outcome units, the
  single-simulator point selector has higher mean held-out decision loss than
  RankCert.
  \item[H4:] Held-out response negative log likelihood is an insufficient
  predictor of tutor-rank stability, operationalized by weak association between
  fit rank and held-out tutor-ranking agreement.
  \item[H5:] RankCert has lower held-out selective risk than a confidence-gated
  single-simulator certificate at comparable coverage, and retains nonzero coverage
  in the stable-world stratum.
\end{description}

\subsection{Uncertainty, Testing, and Multiplicity}

The exact resampling cluster was the ordered pair (held-out family, scenario).
Repeated cohort sizes for the same pair remained in one cluster throughout
resampling and were never treated as independent observations. Clustered bootstrap
intervals resampled these pairs. Superiority and characterization contrasts used
two-sided cluster sign-flip tests, with every cohort-size contribution in a cluster
flipped together. H4 instead used the predeclared two one-sided equivalence
procedure and its corresponding 90\% equivalence interval.

Method contrasts, including H3 and the selective-risk comparison in H5, retained
their setting-level pairing within these clusters. Domain-specific H1 components,
both intersection--union components of H2, the H4 equivalence test, and both
components of H5 entered the same Holm family as the primary H3 test. Both
unadjusted and Holm-adjusted values are reported, and confirmatory interpretation
uses the adjusted family. The materiality rule for H1, the structural-variance
margin for H2, the weak-association bounds for H4, the comparable-coverage tolerance
for H5, bootstrap and randomization counts, and random seeds were applied exactly as
frozen in the development protocol and were not tuned on confirmatory outcomes.

Top-1 agreement and Kendall $\tau_b$ retain ties rather than resolving them with
an ad hoc order. For H4, held-out response negative log likelihood is converted to
fit rank and compared with held-out tutor-ranking agreement using the predeclared
association statistic. For H5, selective risk is compared only at the predeclared
comparable-coverage operating points, and stable-world coverage is evaluated as a
separate component of the same hypothesis.

\subsection{Predeclared Variance Decomposition}

The predeclared decomposition concerns delayed-utility variation, not the paired
decision-loss contrast. Within policy $p$ and cohort size $n$, delayed utility is
decomposed as
\begin{equation}
  U_{pmdr}(n)=\mu_p(n)+\alpha_{pm}(n)+\beta_{pd}(n)
  +(\alpha\beta)_{pmd}(n)+\varepsilon_{pmdr}(n),
  \label{eq:variance-decomposition}
\end{equation}
where $m$ indexes simulator family, $d$ indexes domain, and $r$ indexes residual
parameter and cohort draws. The reported components are simulator family, domain,
family-by-domain, and residual parameter/cohort-draw variation, summarized by
cohort size. H2 uses these predeclared summaries to assess whether the
cohort-draw contribution within the residual term decreases with cohort size while
between-family structural variation remains. No alternative variance decomposition
of the paired RankCert-minus-point loss was introduced. Component estimators and
intervals were those fixed by the development protocol, including boundary
estimates.

\subsection{Missingness, Failures, and Sensitivity Analyses}

Frozen-source and artifact validation passed. The immutable accounting linked all
$1{,}280$ held-out settings to their raw utility chunks, sufficient statistics, and
decision rows. Technical statuses and exclusions were governed by frozen rules
independent of comparative outcomes. Sensitivity analyses were not used to replace
the primary abstention cost, certification thresholds, predictive-adequacy cutoffs,
family weighting, or multiplicity procedure.

\section{Results}

\subsection{Audit, Sample Accounting, and Empirical Calibration}

Frozen-source and artifact validation passed. The confirmatory benchmark contained
$1{,}280$ held-out settings: five held-out oracle families by $64$ scenario draws
per family by four cohort sizes. Each held-out family contributed $16$ scenarios in
each of the stable, ambiguous, structural-conflict, and misspecified strata. The
immutable artifact contained $10{,}240$ decision rows, $163{,}840$ utility
sufficient-statistic rows, and $1{,}280$ raw utility chunks, one per setting.

The confirmatory EdNet-KT1 derivative contained $5{,}000$ learners and $590{,}056$
retained responses in learner-disjoint train, validation, and test partitions.
All five empirical family representatives passed the frozen validation-only
adequacy conjunction before test metrics were examined, so every synthetic fold
contained four visible predictive-adequate families. Table~\ref{tab:calibration}
reports only the held-out test-set metrics; those values did not enter the gate. The
causal-threshold representative is a mechanism-shaped predictive surrogate; these
metrics neither identify nor estimate a causal effect.

\begin{table*}[t]
  \caption{Held-out test-set response metrics for the five calibration-family
  representatives. Each metric used $105{,}074$ test responses. The validation
  gate was locked before these test metrics were examined.}
  \label{tab:calibration}
  \centering
  \begin{tabular}{@{}lrrrrrl@{}}
    \toprule
    Family & AUC & NLL & Brier & ECE & $n$ & Validation gate \\
    \midrule
    BKT-transition & 0.705996 & 0.574345 & 0.195156 & 0.014990 & 105,074 & Pass \\
    PFA-practice & 0.638498 & 0.607420 & 0.209383 & 0.011764 & 105,074 & Pass \\
    Dynamic IRT & 0.730586 & 0.558701 & 0.188768 & 0.014008 & 105,074 & Pass \\
    Attentive history & 0.703100 & 0.575537 & 0.195525 & 0.016120 & 105,074 & Pass \\
    Causal threshold & 0.731104 & 0.558439 & 0.188687 & 0.014546 & 105,074 & Pass \\
    \bottomrule
  \end{tabular}
\end{table*}

\begin{figure}[t]
  \centering
  \IfFileExists{figures/empirical_calibration.pdf}{%
    \includegraphics[width=\linewidth]{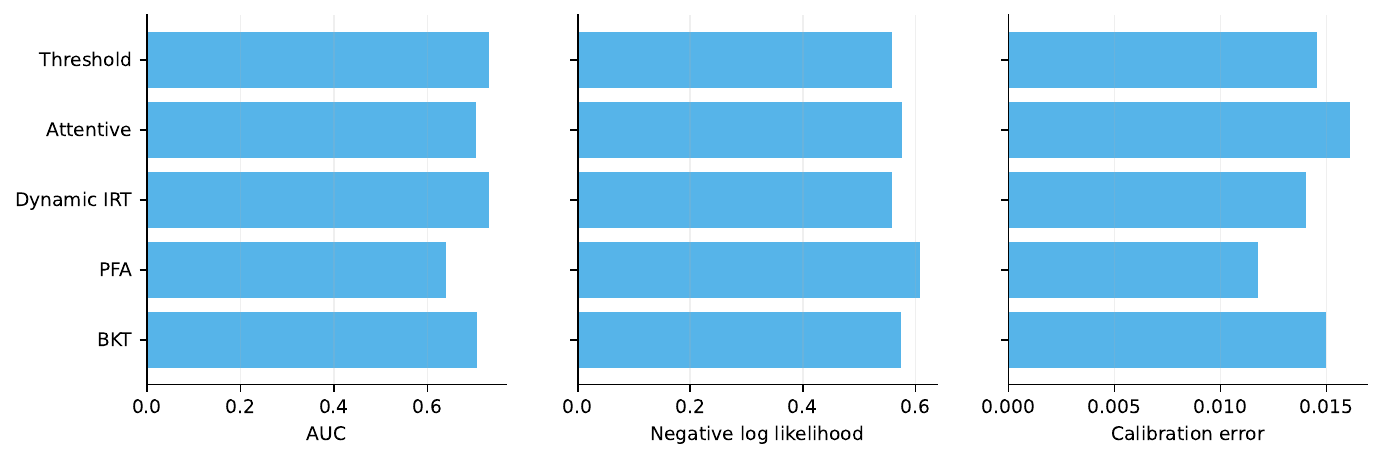}%
  }{%
    \figurefallback{figures/empirical\_calibration.pdf}%
  }
  \caption{Held-out empirical calibration metrics for the five family
  representatives. Adequacy was determined only by the frozen validation gate.}
  \Description{A comparison of held-out AUC, negative log likelihood, Brier score,
  and expected calibration error across the five simulator-family representatives.}
  \label{fig:empirical-calibration}
\end{figure}

\subsection{Confirmatory Hypotheses}

Table~\ref{tab:hypotheses} gives the frozen confirmatory results. H1--H4 were
supported after Holm adjustment. H5 was not supported. The H4 interval is the
predeclared 90\% equivalence interval; the other intervals are 95\% clustered
bootstrap intervals.

\begin{table*}[t]
  \caption{Confirmatory hypothesis results. Effects retain their predeclared
  orientation; H5 is point-certificate minus RankCert selective risk.}
  \label{tab:hypotheses}
  \centering
  \footnotesize
  \begin{tabular}{@{}lp{4.3cm}rp{3.2cm}rrl@{}}
    \toprule
    Hyp. & Estimand & Effect & Interval & Raw $p$ & Holm $p$ & Conclusion \\
    \midrule
    H1 & Minimum-domain mean pairwise top-1 agreement (domain\_b)
       & 0.272917 & 95\% CI $[0.253646,\ 0.293229]$ & 0.000200 & 0.000400 & Supported \\
    H2 & Mean cohort-noise variance at $n=300$ minus $n=30$
       & $-0.000917$ & 95\% CI $[-0.000928,\ -0.000906]$ & 0.000050 & 0.000250 & Supported \\
    H3 & RankCert minus point mean held-out decision loss
       & $-0.006605$ & 95\% CI $[-0.008407,\ -0.004859]$ & 0.000050 & 0.000250 & Supported \\
    H4 & Mean within-setting Kendall $\tau_b$: response-NLL rank versus
       tutor-ranking agreement
       & 0.141644 & 90\% equivalence CI $[0.103632,\ 0.179421]$ & 0.000050 & 0.000250 & Supported \\
    H5 & Point-certificate minus RankCert held-out selective risk
       & $-0.000213$ & 95\% CI $[-0.003238,\ 0.002384]$ & 0.929654 & 0.929654 & Not supported \\
    \bottomrule
  \end{tabular}
\end{table*}

H1 characterized substantial structural disagreement. Mean pairwise top-1
agreement was $0.285417$ in \texttt{domain\_a}, $0.272917$ in
\texttt{domain\_b}, $0.456250$ in \texttt{domain\_c}, and $0.275521$ in
\texttt{domain\_d}. The minimum-domain effect was therefore $0.272917$ (95\% CI
$[0.253646,0.293229]$; raw $p=0.000200$; Holm $p=0.000400$), attained in the
predeclared \texttt{domain\_b}.

Both intersection--union components of H2 were satisfied. Mean cohort-noise
variance at $n=300$ minus $n=30$ was $-0.000917$ (95\% CI
$[-0.000928,-0.000906]$). Between-family structural variance at $n=300$ was
$0.006328$ (95\% CI $[0.006138,0.006523]$), above the frozen $0.0001$
squared-units margin. The composite test had raw $p=0.000050$ and Holm
$p=0.000250$. Larger cohorts reduced within-simulator cohort noise but did not
remove between-family structural variation.

H4 was supported under its equivalence formulation. Mean within-setting Kendall
$\tau_b$ between response-NLL rank and held-out tutor-ranking agreement was
$0.141644$ (90\% equivalence CI $[0.103632,0.179421]$; raw $p=0.000050$;
Holm $p=0.000250$). Thus response fit rank was only weakly associated with
decision-rank stability in this benchmark and was insufficient as a surrogate for
decision validity.

\begin{figure}[t]
  \centering
  \IfFileExists{figures/fit_rank_stability.pdf}{%
    \includegraphics[width=\linewidth]{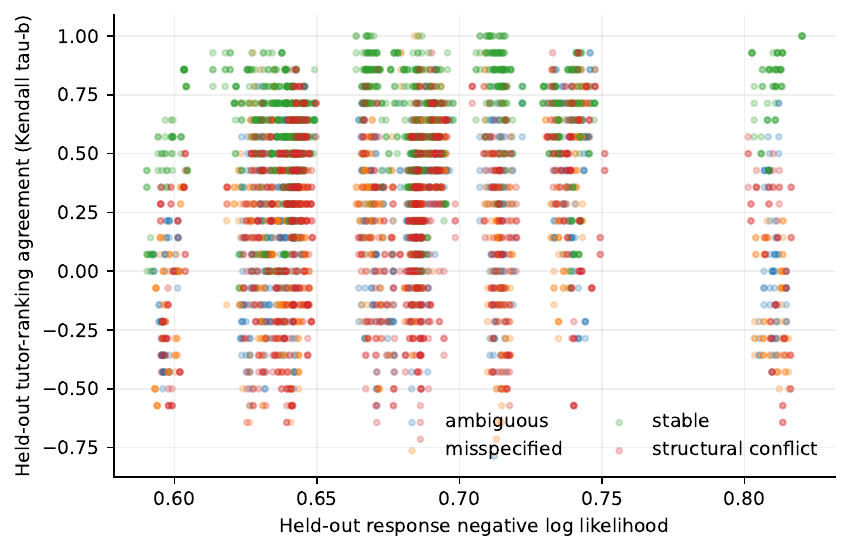}%
  }{%
    \figurefallback{figures/fit\_rank\_stability.pdf}%
  }
  \caption{Association between held-out response-fit rank and held-out
  tutor-ranking agreement. The confirmatory estimand was mean within-setting
  Kendall $\tau_b$.}
  \Description{A plot relating response negative-log-likelihood rank to held-out
  tutor-ranking agreement across confirmatory settings.}
  \label{fig:fit-rank}
\end{figure}

\subsection{Decision Loss}

The primary H3 effect, RankCert minus the full-coverage single-simulator point
selector, was $-0.006605$ normalized-outcome units (95\% CI
$[-0.008407,-0.004859]$; raw $p=0.000050$; Holm $p=0.000250$). Negative values
favor RankCert. This result uses the frozen decision loss in
Eq.~\eqref{eq:decision-loss}: oracle regret after selection and $0.01$ after
abstention. It establishes lower total benchmark decision loss relative to point
selection, not lower regret conditional on selection and not a human-outcome
effect.

\begin{figure}[t]
  \centering
  \IfFileExists{figures/decision_loss_by_cohort.pdf}{%
    \includegraphics[width=\linewidth]{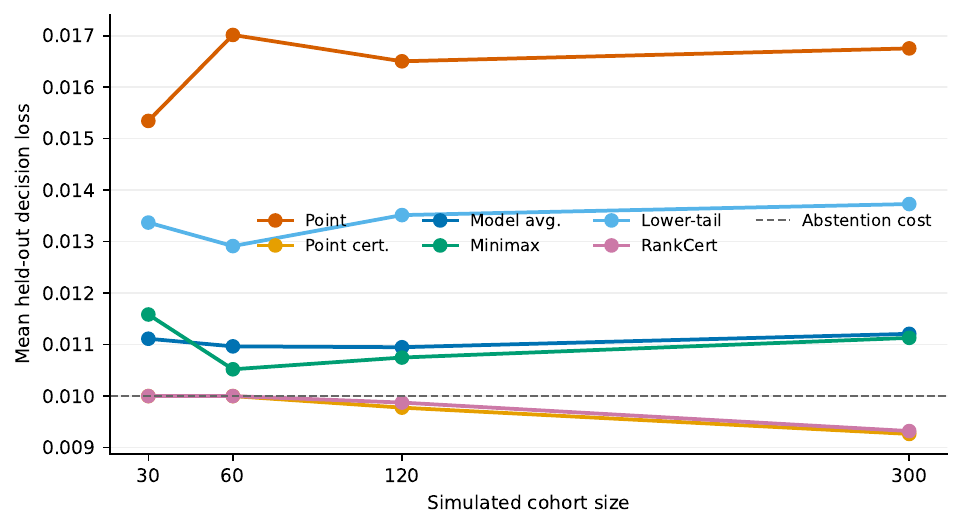}%
  }{%
    \figurefallback{figures/decision\_loss\_by\_cohort.pdf}%
  }
  \caption{Held-out decision loss by cohort size under the frozen abstention cost.
  The confirmatory primary contrast is RankCert minus the full-coverage point
  selector across held-out settings.}
  \Description{Held-out decision loss for RankCert and comparison methods across
  the four simulated cohort sizes.}
  \label{fig:decision-loss}
\end{figure}

\subsection{Coverage and the H5 Null Result}

RankCert certified no policy in any ambiguous, misspecified, or
structural-conflict setting at any cohort size (Table~\ref{tab:coverage}). Stable
settings also had zero coverage at $n=30$ and $n=60$; coverage rose to $0.1125$ at
$n=120$ and $0.4875$ at $n=300$. Because each stratum--cohort cell contained $80$
settings, RankCert made $48$ certifications overall, for aggregate coverage
$0.037500$. These abstentions mean that the evidence failed at least one frozen
certificate condition. They do not show that the policies were ineffective or
equivalent.

\begin{table}[t]
  \caption{RankCert coverage by scenario stratum and cohort size.}
  \label{tab:coverage}
  \centering
  \begin{tabular}{@{}lrrrr@{}}
    \toprule
    Stratum & $n=30$ & $n=60$ & $n=120$ & $n=300$ \\
    \midrule
    Ambiguous & 0.0000 & 0.0000 & 0.0000 & 0.0000 \\
    Misspecified & 0.0000 & 0.0000 & 0.0000 & 0.0000 \\
    Stable & 0.0000 & 0.0000 & 0.1125 & 0.4875 \\
    Structural conflict & 0.0000 & 0.0000 & 0.0000 & 0.0000 \\
    \bottomrule
  \end{tabular}
\end{table}

H5 was not supported. RankCert had coverage $0.037500$, selective risk $0.004627$,
and conditional exact-top error $0.416667$. The confidence-gated point certificate
had coverage $0.042969$, selective risk $0.004414$, and conditional exact-top error
$0.709091$. At comparable overall coverage, with absolute coverage gap $0.00547$,
the point-certificate-minus-RankCert selective-risk difference remained
$-0.000213$ (95\% CI $[-0.003238,0.002384]$; raw $p=0.929654$; Holm
$p=0.929654$). The interval includes zero, and the negative point estimate is not
in the direction required to show lower RankCert selective risk.

Conditional exact-top error treats every selected policy outside the oracle's
exact top set as wrong, even when its oracle regret is tiny. It is therefore not
interchangeable with selective risk, which retains regret magnitude. RankCert did
retain nonzero coverage in the stable-world stratum, but that component did not
rescue the composite H5 claim. The supported advantage is lower total decision
loss relative to full-coverage point selection, not lower conditional regret than
the confidence-gated point certificate.

\begin{figure}[t]
  \centering
  \IfFileExists{figures/risk_coverage.pdf}{%
    \includegraphics[width=\linewidth]{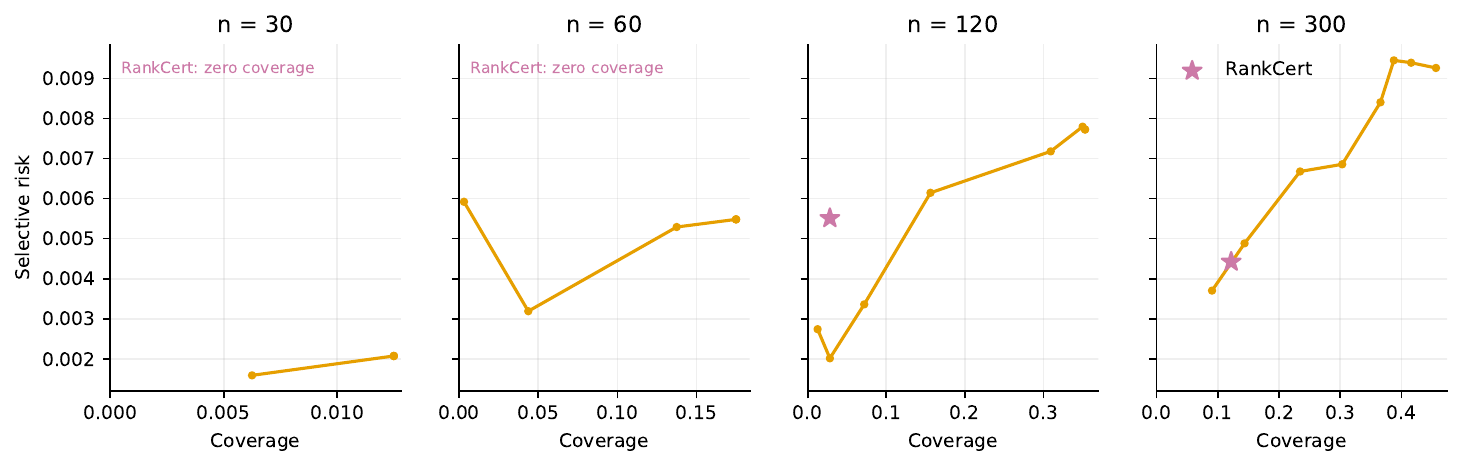}%
  }{%
    \figurefallback{figures/risk\_coverage.pdf}%
  }
  \caption{Selective risk and coverage for RankCert and the confidence-gated
  point certificate. H5 compared selective risk only at the predeclared comparable
  coverage operating points.}
  \Description{Risk--coverage operating points for RankCert and the
  confidence-gated single-simulator point certificate.}
  \label{fig:risk-coverage}
\end{figure}

\subsection{Certificate Gate Diagnostics}

These gate diagnostics were descriptive and post-confirmatory. RankCert
certification is a conjunction, so marginal pass rates do not assign causal
responsibility for selection or abstention to any individual gate. Multiple gates
can and did fail in the same setting. The unique-candidate and
predictive-adequacy gates each passed in all $1{,}280$ settings; Table~\ref{tab:gate-diagnostics}
reports the remaining marginal pass rates and final certification rate.

\begin{table}[t]
  \caption{Descriptive marginal certificate-gate pass rates by scenario stratum.
  LODO and LOFO denote leave-one-domain-out and
  leave-one-visible-family-out average survival.}
  \label{tab:gate-diagnostics}
  \centering
  \small
  \begin{tabular}{@{}lrrrrrr@{}}
    \toprule
    Stratum & $P(\mathrm{best})$ & $q_{.95}$ regret & Domain support & LODO & LOFO & Certified \\
    \midrule
    Ambiguous & 0.000000 & 0.303125 & 0.290625 & 0.393750 & 0.050000 & 0.000000 \\
    Misspecified & 0.000000 & 0.328125 & 0.593750 & 0.650000 & 0.134375 & 0.000000 \\
    Stable & 0.165625 & 0.484375 & 0.871875 & 0.900000 & 0.662500 & 0.150000 \\
    Structural conflict & 0.000000 & 0.190625 & 0.293750 & 0.340625 & 0.087500 & 0.000000 \\
    Overall & 0.041406 & 0.326562 & 0.512500 & 0.571094 & 0.233594 & 0.037500 \\
    \bottomrule
  \end{tabular}
\end{table}

The probability-best gate passed in no ambiguous, misspecified, or
structural-conflict setting, consistent with zero certification in those strata.
Marginally, probability-best and leave-one-visible-family-out survival had the
lowest overall pass rates among the reported gates. This comparison is descriptive:
because gates co-failed, it does not show that either gate singly caused an
abstention.

\subsection{Delayed-Utility Variance Decomposition}

Table~\ref{tab:variance} reports mean variance shares by cohort size. The residual
share declined from $0.144043$ at $n=30$ to $0.090389$ at $n=300$, whereas the
family share remained nonzero and increased from $0.274654$ to $0.291726$. Domain
variation was the largest component at every cohort size, and the
family-by-domain share remained approximately stable. These shares complement the
H2 raw-variance contrast; they are not a decomposition of the H3 decision-loss
effect.

\begin{table}[t]
  \caption{Mean delayed-utility variance shares by cohort size. Rows sum to one up
  to displayed precision.}
  \label{tab:variance}
  \centering
  \begin{tabular}{@{}rrrrr@{}}
    \toprule
    Cohort & Family & Domain & Family $\times$ domain & Residual \\
    \midrule
    30 & 0.274654 & 0.544892 & 0.036412 & 0.144043 \\
    60 & 0.284016 & 0.563664 & 0.037155 & 0.115165 \\
    120 & 0.288655 & 0.573649 & 0.037518 & 0.100178 \\
    300 & 0.291726 & 0.580162 & 0.037723 & 0.090389 \\
    \bottomrule
  \end{tabular}
\end{table}

\begin{figure}[t]
  \centering
  \IfFileExists{figures/variance_decomposition.pdf}{%
    \includegraphics[width=\linewidth]{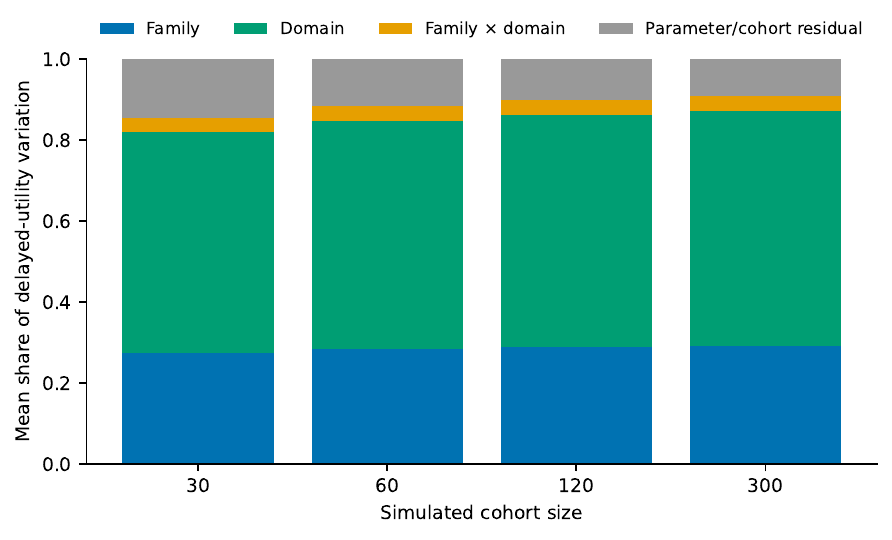}%
  }{%
    \figurefallback{figures/variance\_decomposition.pdf}%
  }
  \caption{Mean delayed-utility variance shares attributable to simulator family,
  domain, family-by-domain interaction, and residual parameter/cohort draws.}
  \Description{Stacked variance shares for simulator family, domain,
  family-by-domain interaction, and residual variation at each cohort size.}
  \label{fig:variance}
\end{figure}

\subsection{Sensitivity Analyses}

The analyses in this subsection were secondary and post-confirmatory. They were not
used to tune the frozen primary thresholds, abstention cost, family weights,
actions, or estimands. First, immutable-chunk replay reproduced the frozen RankCert
action in all $1{,}280$ settings.

Table~\ref{tab:abstention-sensitivity} re-evaluates the RankCert-minus-point mean
decision-loss contrast under alternative abstention costs. Negative values favor
RankCert. The ranking reversed between costs $0.01$ and $0.02$; the H3 advantage
therefore should not be described as robust above the frozen $0.01$ cost.

\begin{table}[t]
  \caption{Secondary abstention-cost sensitivity for the RankCert-minus-point mean
  held-out decision-loss contrast.}
  \label{tab:abstention-sensitivity}
  \centering
  \begin{tabular}{@{}rr@{}}
    \toprule
    Abstention cost & Mean loss contrast \\
    \midrule
    0.0000 & $-0.016230$ \\
    0.0025 & $-0.013824$ \\
    0.0050 & $-0.011418$ \\
    0.0100 & $-0.006605$ \\
    0.0200 & $+0.003020$ \\
    0.0500 & $+0.031895$ \\
    \bottomrule
  \end{tabular}
\end{table}

A threshold grid crossed probability-best thresholds $0.40$, $0.50$, and $0.60$
with 95th-percentile regret thresholds $0.03$, $0.05$, and $0.07$. At abstention
cost $0.01$, RankCert coverage ranged from $0.007031$ to $0.103906$, and mean
decision loss ranged from $0.009444$ to $0.009965$. The frozen point selector mean
loss was $0.016404$. Replaying the primary thresholds produced coverage $0.037500$,
mean decision loss $0.009799$, and selective risk $0.004627$.

Modest family-weight perturbations duplicated one visible family's draws, changing
that family's mass from $1/4$ to $2/5$. Across the five perturbations, coverage
ranged from $0.028125$ to $0.063281$, selective risk from $0.003254$ to $0.007219$,
and mean decision loss from $0.009584$ to $0.009877$. Equal weighting produced
coverage $0.037500$, selective risk $0.004627$, and mean decision loss $0.009799$.
These replays are robustness checks, not alternative primary estimands.

\section{Discussion}

\subsection{Lower Total Decision Loss, but No Selective-Risk Advantage}

RankCert lowered overall held-out decision loss relative to the full-coverage
single-simulator point selector. Under the frozen loss, this advantage reflects the
combined consequences of the choices RankCert made and its abstentions, each of
which cost $0.01$. It therefore supports the decision-level H3 claim inside the
benchmark. It does not show that RankCert's selected policies had lower conditional
regret than every selective alternative.

The post-confirmatory cost replay also limits H3. The RankCert-minus-point contrast
was negative at the frozen $0.01$ cost but became positive at $0.02$; the ranking
reversed between those values. The lower total decision loss is therefore a result
under the declared abstention cost, not a cost-insensitive ordering.

The H5 null makes that distinction concrete. At comparable coverage, RankCert did
not have lower selective risk than the confidence-gated point certificate. The
point estimate slightly favored the point certificate under the stated orientation,
the interval spanned zero, and the Holm-adjusted $p$-value was $0.929654$. The
stable-world coverage component was nonzero, but the composite hypothesis was not
supported. RankCert's empirical advantage is consequently lower total decision
loss relative to full-coverage point selection, not lower conditional regret than
the comparable confidence-gated certificate.

\subsection{Structural Uncertainty Was Decision-Relevant}

H1 showed that top-1 tutor-policy agreement was far from perfect in the weakest
predeclared domain. H2 separated the effect of cohort size from structural
variation: cohort noise decreased as the simulated cohort grew, yet the family
variance share remained nonzero. H4 further showed that response-NLL rank was only
weakly associated with held-out tutor-ranking agreement. Predictive adequacy was
still necessary for model inclusion, but relative predictive fit was not a valid
substitute for evaluating decision-rank stability. This pattern is consistent with
the broader concern that predictively credible models can remain consequentially
underspecified for downstream decisions \cite{damour2022}.

\subsection{Abstention Marks a Boundary of Evidence}

The empirical answer to ``when'' was narrow: certification occurred only in stable
scenarios at $n=120$ or $n=300$, and only when the full certificate conjunction
held. RankCert certified no policy in ambiguous, misspecified, or
structural-conflict settings, and it also abstained in all stable settings at
$n=30$ and $n=60$. Marginally, probability-best and
leave-one-visible-family-out survival were particularly restrictive, with overall
pass rates $0.041406$ and $0.233594$, respectively. Those diagnostics are
descriptive, not causal, because gates co-failed and certification required every
gate. The outcomes show that the benchmark evidence did not justify a single robust
choice; they do not show that any policy was ineffective, that policies were equal,
or that a human learner would fail to benefit.

Low aggregate coverage also limits practical interpretation. A procedure that
acts in $0.037500$ of held-out settings delegates most cases to abstention. Whether
that tradeoff is appropriate depends on the benchmark-specific cost of abstention
and on what a downstream system does after receiving
\emph{\AbstentionMessage} The certificate should not be converted into a fallback
recommendation that bypasses its own evidence boundary.

\subsection{Scope of ``Safe''}

RankCert is safe only in the restricted, benchmark-scoped sense defined here: a
selection passed the frozen gates under the declared utility and uncertainty set.
The experiment did not measure human learning, validate the simulated utility
against people, test a deployed tutor, estimate a causal treatment effect, or
certify fairness, privacy, security, accessibility, or general safety. In
particular, the causal-threshold family is a mechanism-shaped predictive surrogate,
not a causal estimator.

\section{Threats to Validity}

\textbf{Fixed simulator inventory.} The five families are structurally varied but
cannot span every plausible learner mechanism. Shared misspecification can produce
agreement without truth, and leave-one-family-out evaluation tests only the
declared forms and perturbations.

\textbf{Hand-designed policies and utility.} The eight policy profiles,
equal-budget constraint, delayed-utility scale, oracle regret, and abstention cost
were designed for this benchmark. Their value judgments need not transfer to other
tutoring objectives. The $0.01$ abstention cost is especially consequential when
policy regrets are small: the secondary replay reversed the RankCert--point ordering
between costs $0.01$ and $0.02$. H3 therefore depends on the declared cost and does
not establish robustness above $0.01$.

\textbf{Public secondary-data calibration.} The licensed, de-identified EdNet-KT1
derivative constrained empirical plausibility but may not represent other domains,
platforms, learners, or interventions. Calibration adequacy is not human-outcome
validation.

\textbf{Synthetic held-out oracle.} Oracle regret is observable because the
evaluation environments are synthetic. A real deployment does not provide the same
oracle, and the benchmark cannot establish proxy-to-human validity.

\textbf{Benchmark-specific thresholds.} Probability-best, regret, domain-rank,
family-count, survival, and adequacy thresholds were frozen before analysis, but
their appropriateness is conditional on this utility and scenario construction.

\textbf{Low coverage.} RankCert certified only $0.037500$ of settings overall and no
ambiguous, misspecified, or structural-conflict setting. This sharply limits any
claim about how often the method can provide a decision.

\textbf{No human validation.} No participant was recruited and no tutoring policy
was delivered to a person. The study cannot estimate human learning, causal effects,
disparate impact, acceptability, teacher workload, accessibility, deployment
effectiveness, or general safety.

\textbf{Statistical and computational validity.} Monte Carlo error, posterior
approximation, optimizer behavior, and ties can affect method contrasts. Frozen
seeds, immutable manifests, paired analyses, clustered intervals, and Holm
adjustment reduce but do not eliminate these risks.

\section{Ethics and Responsible Use}

This study recruited no participants, interacted with no learner, and administered
no intervention. It used licensed, de-identified secondary data for calibration and
synthetic environments for decision evaluation. Processing was limited to retained
fields required by the calibration, and no attempt was made to re-identify
individuals. No ethics approval or exemption is claimed in this article.

RankCert should not be presented as a general safety certificate. Its output is
conditional on candidate policies, utility, data, predictive checks, structural
families, and scenario design. In particular, certification does not establish
fairness, privacy, security, accessibility, or benefit for human learners. An
abstention must remain visible as \emph{\AbstentionMessage} and must not be converted
into an implicit recommendation. A human-facing study would require a separate
protocol, appropriate review, and outcome measures suited to people.

\section{Reproducibility and Artifact Availability}

The artifact accompanying the submission uses a local tree rooted at
\path{main.tex}, with the verified bibliography source at \path{references.bib}
and rendered figures under \path{figures/}. This describes the accompanying
artifact and does not assert a public URL. The expected figure paths are:
\begin{itemize}
  \item \path{figures/decision_loss_by_cohort.pdf};
  \item \path{figures/risk_coverage.pdf};
  \item \path{figures/variance_decomposition.pdf};
  \item \path{figures/empirical_calibration.pdf}; and
  \item \path{figures/fit_rank_stability.pdf}.
\end{itemize}
Conditional inclusion preserves a compilable manuscript if a figure file is absent
from a working copy.

The immutable confirmatory manifest records frozen-source validation, configuration
and seed provenance, five oracle-family folds, $1{,}280$ raw utility chunks,
$163{,}840$ utility sufficient-statistic rows, $10{,}240$ decision rows, and the
analysis outputs used for the tables and figures. Development and pilot outputs are
separated from confirmatory outputs and are not manuscript findings. The licensed
EdNet-KT1 data are not redistributed; the artifact records derivative construction
and learner-disjoint partitioning without exposing source records.

The post-confirmatory replay and sensitivity records are stored at:
\begin{itemize}
  \item \path{confirmatory/sensitivity/abstention_cost.csv};
  \item \path{confirmatory/sensitivity/certificate_replay.csv};
  \item \path{confirmatory/sensitivity/certificate_replay_summary.csv};
  \item \path{confirmatory/sensitivity/metadata.json};
  \item \path{confirmatory/sensitivity/gate_diagnostics.csv}; and
  \item \path{confirmatory/sensitivity/gate_pass_rates.csv}.
\end{itemize}
The replay files link immutable utility chunks to recomputed certificate actions;
the metadata records the sensitivity grid and family-weight perturbations. The gate
diagnostic files contain setting-level gate outcomes and the aggregated marginal
pass rates reported in Table~\ref{tab:gate-diagnostics}.

\section{Conclusion}

RankCert is an abstaining decision procedure for choosing among
equal-budget AI tutoring policies under structural learner-model uncertainty. Its
certificate requires agreement across complementary posterior, regret,
cross-domain, predictive-adequacy, leave-one-domain-out-average, and
leave-one-visible-family-out-average checks. Across $1{,}280$ held-out settings,
H1, H2, H3, and H4 were supported: simulator families often disagreed on the top
policy, larger cohorts reduced cohort noise without eliminating structural
variance, RankCert lowered total decision loss relative to full-coverage point
selection at the frozen $0.01$ abstention cost, and response-fit rank was only
weakly associated with tutor-rank stability. H5 was not supported. RankCert did
not lower selective risk relative to the comparable confidence-gated point
certificate, although it retained nonzero coverage in stable settings. Its frequent
abstention, including zero certification
in all ambiguous, misspecified, and structural-conflict settings, marks a boundary
of benchmark evidence rather than a finding about tutor efficacy. These conclusions
apply only to benchmark-scoped decision certification under the declared utility
and uncertainty set; they do not establish human learning, causal effects,
deployment effectiveness, or general safety.

\begin{acks}
The author received no specific funding for this work.
OpenAI Prism was used during manuscript preparation to support drafting,
language refinement, code generation, and analysis-workflow orchestration. The
author designed the study, fixed the protocol and decision definitions, executed
and reviewed the analyses, verified every reported result against the frozen
artifacts, and assumes full responsibility for the content.
\end{acks}

\bibliographystyle{ACM-Reference-Format}
\bibliography{references}

@article{corbett1995,
  author = {Corbett, Albert T. and Anderson, John R.},
  title = {Knowledge Tracing: Modeling the Acquisition of Procedural Knowledge},
  journal = {User Modeling and User-Adapted Interaction},
  year = {1995},
  volume = {4},
  number = {4},
  pages = {253--278},
  doi = {10.1007/BF01099821}
}

@inproceedings{pavlik2009,
  author = {Pavlik, Philip I. and Cen, Hao and Koedinger, Kenneth R.},
  title = {Performance Factors Analysis: A New Alternative to Knowledge Tracing},
  booktitle = {Artificial Intelligence in Education},
  year = {2009},
  pages = {531--538},
  doi = {10.3233/978-1-60750-028-5-531}
}

@inproceedings{piech2015,
  author = {Piech, Chris and Bassen, Jonathan and Huang, Jonathan and Ganguli, Surya and Sahami, Mehran and Guibas, Leonidas and Sohl-Dickstein, Jascha},
  title = {Deep Knowledge Tracing},
  booktitle = {Advances in Neural Information Processing Systems 28},
  year = {2015},
  pages = {505--513},
  url = {https://proceedings.neurips.cc/paper/2015/hash/bac9162b47c56fc8a4d2a519803d51b3-Abstract.html}
}

@inproceedings{ghosh2020,
  author = {Ghosh, Aritra and Heffernan, Neil and Lan, Andrew S.},
  title = {Context-Aware Attentive Knowledge Tracing},
  booktitle = {Proceedings of the 26th ACM SIGKDD International Conference on Knowledge Discovery and Data Mining},
  year = {2020},
  pages = {2330--2339},
  doi = {10.1145/3394486.3403282}
}

@article{abdelrahman2023,
  author = {Abdelrahman, Ghodai and Wang, Qing and Nunes, Bernardo},
  title = {Knowledge Tracing: A Survey},
  journal = {ACM Computing Surveys},
  year = {2023},
  volume = {55},
  number = {11},
  pages = {1--37},
  doi = {10.1145/3569576}
}

@inproceedings{liu2022,
  author = {Liu, Zitao and Liu, Qiongqiong and Chen, Jiahao and Huang, Shuyan and Tang, Jiliang and Luo, Weiqi},
  title = {{pyKT}: A Python Library to Benchmark Deep Learning Based Knowledge Tracing Models},
  booktitle = {Advances in Neural Information Processing Systems 35: Datasets and Benchmarks Track},
  year = {2022},
  url = {https://proceedings.neurips.cc/paper_files/paper/2022/hash/75ca2b23d9794f02a92449af65a57556-Abstract-Datasets_and_Benchmarks.html}
}

@article{choi2020,
  author = {Choi, Youngduck and Lee, Youngnam and Shin, Dongmin and Cho, Junghyun and Park, Seoyon and Lee, Seewoo and Baek, Jineon and Bae, Chan and Kim, Byungsoo and Heo, Jaewe},
  title = {{EdNet}: A Large-Scale Hierarchical Dataset in Education},
  journal = {arXiv preprint arXiv:1912.03072},
  year = {2020},
  eprint = {1912.03072},
  archivePrefix = {arXiv},
  url = {https://arxiv.org/abs/1912.03072}
}

@article{damour2022,
  author = {D'Amour, Alexander and Heller, Katherine and Moldovan, Dan and others},
  title = {Underspecification Presents Challenges for Credibility in Modern Machine Learning},
  journal = {Journal of Machine Learning Research},
  year = {2022},
  volume = {23},
  number = {226},
  pages = {1--61},
  url = {https://www.jmlr.org/papers/v23/20-1335.html}
}

@article{breiman2001,
  author = {Breiman, Leo},
  title = {Statistical Modeling: The Two Cultures},
  journal = {Statistical Science},
  year = {2001},
  volume = {16},
  number = {3},
  pages = {199--231},
  doi = {10.1214/ss/1009213726}
}

@article{nilim2005,
  author = {Nilim, Arnab and El Ghaoui, Laurent},
  title = {Robust Control of Markov Decision Processes with Uncertain Transition Matrices},
  journal = {Operations Research},
  year = {2005},
  volume = {53},
  number = {5},
  pages = {780--798},
  doi = {10.1287/opre.1050.0216}
}

@article{hansen2007,
  author = {Hansen, Bruce E.},
  title = {Least Squares Model Averaging},
  journal = {Econometrica},
  year = {2007},
  volume = {75},
  number = {4},
  pages = {1175--1189},
  doi = {10.1111/j.1468-0262.2007.00785.x}
}

@inproceedings{geifman2017,
  author = {Geifman, Yonatan and El-Yaniv, Ran},
  title = {Selective Classification for Deep Neural Networks},
  booktitle = {Advances in Neural Information Processing Systems 30},
  year = {2017},
  pages = {4878--4887},
  url = {https://proceedings.neurips.cc/paper/2017/hash/4a8423d5e91fda00bb7e46540e2b0cf1-Abstract.html}
}

@inproceedings{geifman2019,
  author = {Geifman, Yonatan and El-Yaniv, Ran},
  title = {{SelectiveNet}: A Deep Neural Network with an Integrated Reject Option},
  booktitle = {Proceedings of the 36th International Conference on Machine Learning},
  year = {2019},
  volume = {97},
  series = {Proceedings of Machine Learning Research},
  pages = {2151--2159},
  url = {https://proceedings.mlr.press/v97/geifman19a.html}
}

@article{doroudi2019,
  author = {Doroudi, Shayan and Aleven, Vincent and Brunskill, Emma},
  title = {Where's the Reward? A Review of Reinforcement Learning for Instructional Sequencing},
  journal = {International Journal of Artificial Intelligence in Education},
  year = {2019},
  volume = {29},
  number = {4},
  pages = {568--620},
  doi = {10.1007/s40593-019-00187-x}
}

@inproceedings{mandel2014,
  author = {Mandel, Travis and Liu, Yun-En and Levine, Sergey and Brunskill, Emma and Popovi{\'c}, Zoran},
  title = {Offline Policy Evaluation Across Representations with Applications to Educational Games},
  booktitle = {Proceedings of the 13th International Conference on Autonomous Agents and Multiagent Systems},
  year = {2014},
  pages = {1077--1084},
  publisher = {International Foundation for Autonomous Agents and Multiagent Systems},
  url = {https://grail.cs.washington.edu/projects/ordering/}
}

@inproceedings{hoiles2016,
  author = {Hoiles, William and van der Schaar, Mihaela},
  title = {Bounded Off-Policy Evaluation with Missing Data for Course Recommendation and Curriculum Design},
  booktitle = {Proceedings of the 33rd International Conference on Machine Learning},
  series = {Proceedings of Machine Learning Research},
  volume = {48},
  pages = {1596--1604},
  year = {2016},
  publisher = {PMLR},
  url = {https://proceedings.mlr.press/v48/hoiles16.html}
}

@article{wilder2019,
  author = {Wilder, Bryan and Dilkina, Bistra and Tambe, Milind},
  title = {Melding the Data-Decisions Pipeline: Decision-Focused Learning for Combinatorial Optimization},
  journal = {Proceedings of the AAAI Conference on Artificial Intelligence},
  year = {2019},
  volume = {33},
  number = {1},
  pages = {1658--1665},
  doi = {10.1609/aaai.v33i01.33011658}
}

@article{elmachtoub2022,
  author = {Elmachtoub, Adam N. and Grigas, Paul},
  title = {Smart ``Predict, then Optimize''},
  journal = {Management Science},
  year = {2022},
  volume = {68},
  number = {1},
  pages = {9--26},
  doi = {10.1287/mnsc.2020.3922}
}

@article{kennedy2001,
  author = {Kennedy, Marc C. and O'Hagan, Anthony},
  title = {Bayesian Calibration of Computer Models},
  journal = {Journal of the Royal Statistical Society: Series B (Statistical Methodology)},
  year = {2001},
  volume = {63},
  number = {3},
  pages = {425--464},
  doi = {10.1111/1467-9868.00294}
}

\end{document}